\documentclass[lettersize,journal]{IEEEtran}
\usepackage{amsmath,amsfonts,amssymb}
\usepackage{array}
\usepackage{multirow}
\usepackage{booktabs}
\usepackage[caption=false,font=normalsize,labelfont=sf,textfont=sf]{subfig}
\usepackage{textcomp}
\usepackage{stfloats}
\usepackage{url}
\usepackage{graphicx}
\usepackage{xcolor}
\usepackage{float}
\usepackage{bbding}
\usepackage{cite}
\usepackage[hidelinks]{hyperref}
\begin{document}
\title{Solar-Conditioned Flow Matching for Multitemporal\\
Aerial Visible-to-Infrared Image Translation}
\author{Wenwen~Fang, Bingwei~Hui, Xin~Chen, Zhongzhen~Sun, and~Zhenfeng~Shao%
\thanks{This work was supported in part by the National Natural Science
Foundation of China under Grant U24B20189. This work has been submitted to
the IEEE Geoscience and Remote Sensing Letters for possible publication.
(Corresponding author: Bingwei Hui.)}%
\thanks{Wenwen~Fang, Bingwei~Hui, Xin~Chen, and~Zhongzhen~Sun are with the National Key Laboratory of Automatic Target Recognition, National University of Defense Technology, Changsha 410073, China (e-mail: huibingwei07@nudt.edu.cn).}%
\thanks{Zhenfeng Shao is with the State Key Laboratory of Information Engineering in Surveying, Mapping and Remote Sensing, Wuhan University, Wuhan 430079, China (e-mail: shaozhenfeng@whu.edu.cn).}}
\markboth{}{}
\maketitle
\begin{abstract}
Aerial visible-to-infrared (V2IR) image translation provides scalable
thermal data across the full diurnal cycle. However, existing
methods predominantly learn a static mapping within narrow acquisition windows.
As the thermal state evolves throughout the day, a single visible image may
correspond to multiple plausible IR appearances, making V2IR translation an
inherently multitemporal task. This
letter proposes solar-conditioned latent flow matching (Solar-Flow), which
upgrades aerial V2IR translation from static
mapping to multitemporal generation. First, we design three computable
physical factors, namely the solar elevation, the thermal phase clock (TPC), and
the flight altitude.
The TPC counts the elapsed heating and cooling hours on a continuous
24-hour axis anchored at sunrise, so the thermal state stays well defined
even at hours without training samples. Second, we inject the condition into every Transformer block of a pretrained
flow matching generator through adaptive layer normalization. This design
preserves translation fidelity and enables multitemporal generation.
Finally, we construct a paired aerial V2IR dataset covering 14 hour groups.
Solar-Flow achieves multitemporal generation on this dataset, where a
single visible image is translated into plausible IR images at multiple
times of day, even at hours that were never seen in training.
\end{abstract}
\section{Introduction}
\IEEEPARstart{T}{hermal} infrared (IR) imagery is essential for Earth
observation because it captures the radiation emitted by the scene itself and
remains informative at night and under weak illumination. However, constructing
large paired visible-to-IR (V2IR) datasets requires specialized acquisition
hardware and pixel-level registration, making aerial V2IR translation an
urgent research direction \cite{ustnet,dravit}. Image-to-image
translation between other sensor pairs such as SAR and optical imagery has
also been extensively studied \cite{dusar,baicd}, yet these approaches treat
the mapping as static and time agnostic, ignoring thermal state evolution.
Recent generative approaches such as latent diffusion \cite{ldm} and
ThermalGen \cite{thermalgen} produce realistic thermal appearances but remain
blind to the underlying thermal state. As a result, one visible input always
yields a single fixed IR appearance. PID \cite{pid} and IPLD \cite{ipld}
embed thermodynamic decomposition into IR synthesis, yet generation is never
conditioned on the thermal state, so each
remains a static mapping. Surfaces accumulate heat after sunrise and release it
after sunset, so the same scene exhibits different IR appearances at different
times of the day. Moreover, existing aerial benchmarks such as AVIID
\cite{aviid} are recorded within short acquisition windows, so this temporally
varying behavior cannot be learned. To the best of our knowledge, a paired
aerial V2IR dataset covering the full diurnal cycle is still absent.
To address these limitations, we propose a multitemporal V2IR framework
that shifts the generative objective from static mapping to multitemporal
generation. We also construct a paired aerial V2IR
dataset covering the full diurnal cycle.
Specifically, each frame is paired with three physical factors computed from
its timestamp and GPS coordinate alone, namely the solar elevation, the thermal
phase clock (TPC), and the flight altitude. The TPC measures the elapsed heating
hours during the day and the elapsed cooling hours after sunset within a
continuous 24-hour coordinate. This provides a physically meaningful time axis
that can be compared across dates. The condition is injected into ThermalGen
through adaptive layer normalization. This allows a single visible image to be
translated into IR images at any time, including hours unseen in training. To
build this dataset, daytime pairs come from synchronized dual-camera videos,
and night pairs from georeferenced orthophotos. The main contributions are as
follows.
\begin{enumerate}
\item We propose solar-conditioned latent flow matching (Solar-Flow), which
upgrades aerial V2IR translation from static mapping to multitemporal
generation. The
physical condition is injected into every Transformer
block of a pretrained latent flow matching generator through adaptive layer
normalization. This design unifies translation fidelity and multitemporal
controllability in a single model.
\item We design a computable physical condition composed of three factors,
namely the thermal phase clock (TPC), the solar elevation, and the flight
altitude.
Anchored at sunrise on a continuous 24-hour axis, the TPC keeps the thermal
state well defined even at hours without training samples. The same condition
transfers to GAN-based translators, granting them multitemporal control.
\item We construct the first paired aerial V2IR dataset with full diurnal
coverage. Daytime pairs are captured by synchronized dual cameras and
registered with a calibrated homography, while nighttime pairs are obtained by
projecting a georeferenced orthophoto onto each thermal frame.
\end{enumerate}
\section{Method}
\label{sec:method}
\begin{figure*}[!t]
\centering
\includegraphics[width=0.9\textwidth]{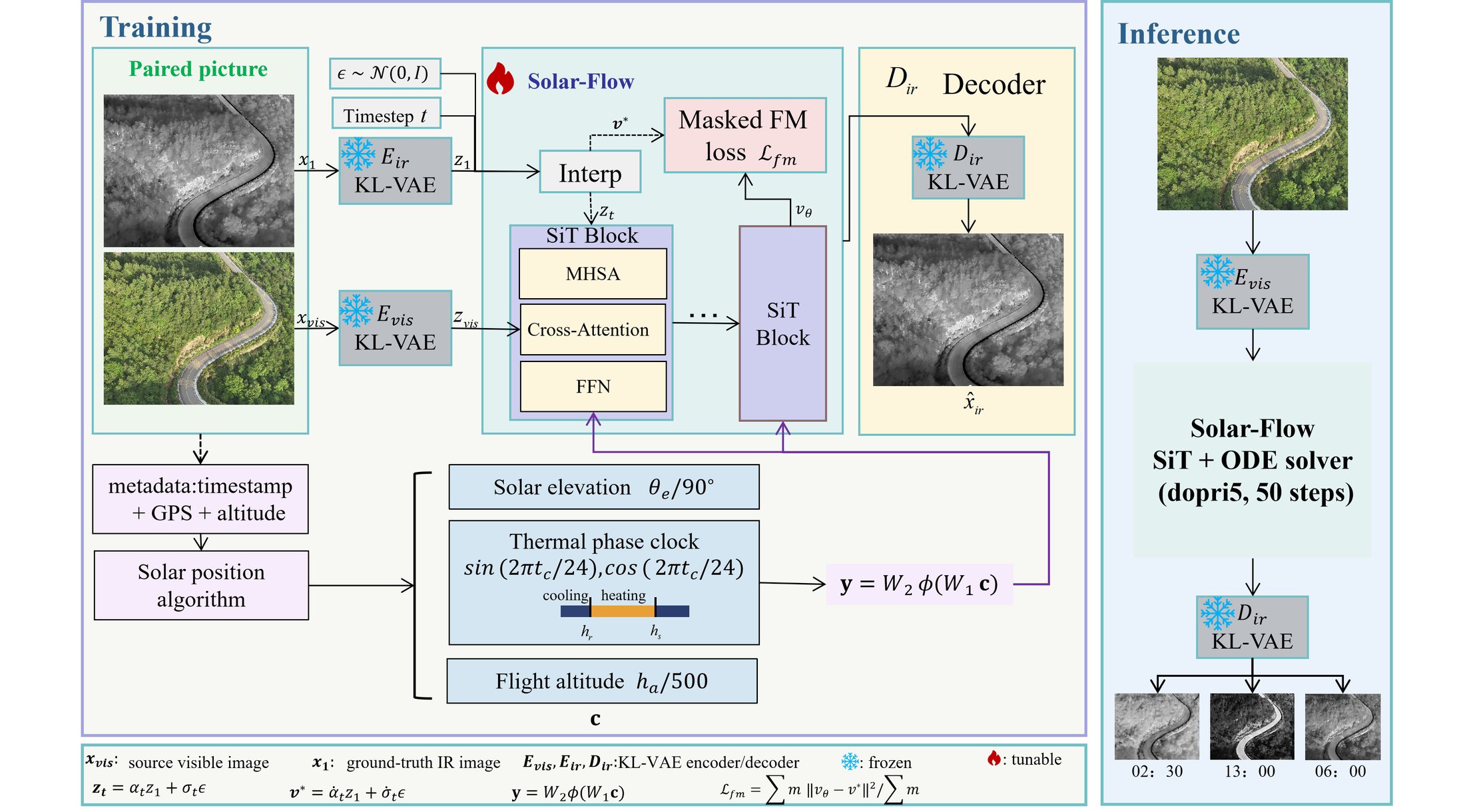}
\caption{Overall framework of the proposed method. The core design upgrades a
pretrained flow matching generator from static mapping to
multitemporal generation, where a computable physical condition is injected into every
Transformer block through adaptive layer normalization, and the source image
tokens are fused via cross-attention.}
\label{fig:framework}
\end{figure*}

The overall framework of Solar-Flow is illustrated
in Fig.~\ref{fig:framework}. The model takes a source image and the physical condition as
inputs and outputs the corresponding IR image.
We first review the flow matching backbone in Section~\ref{sec:backbone},
then detail the three physical factors in Section~\ref{sec:cond}, and finally
describe how the condition is injected and the model is trained in
Section~\ref{sec:scfm}.
\subsection{Flow Matching Backbone}
\label{sec:backbone}
Aerial V2IR translation operates on high-resolution imagery. Training a
large generator from scratch on a single-site dataset is therefore
impractical. We build upon ThermalGen \cite{thermalgen}, a
pretrained latent flow matching model for RGB to thermal translation, and
replace its original label condition with the proposed physical condition.
Two frozen KL-VAEs \cite{ldm} bridge the pixel
and latent spaces. The visible one encodes the source image into a
conditional latent $z_{\mathrm{vis}}$, while the thermal one maps the ground
truth IR image to the target latent $z_1$. Each $640\times512$ image pair is
thereby compressed to an $80\times64$ latent pair, and each latent is then
processed as 1280 tokens. An interpolant Transformer (SiT)
\cite{sit,dit} with the XL/2 configuration then learns the velocity field
that transports a Gaussian prior to the target latent $z_1$. Following the
interpolation schedule of \cite{sit}, we define the intermediate state and
its target velocity as
\begin{equation}
z_t = \alpha_t z_1 + \sigma_t \epsilon, \qquad \epsilon \sim \mathcal{N}(0, I)
\label{eq:interp}
\end{equation}
\begin{equation}
v^{*}(z_t, t) = \dot{\alpha}_t z_1 + \dot{\sigma}_t \epsilon
\label{eq:target}
\end{equation}
where $\alpha_t$ and $\sigma_t$ satisfy the boundary conditions
$\alpha_0{=}1$, $\sigma_0{=}0$, $\alpha_1{=}0$, $\sigma_1{=}1$. As in
ThermalGen \cite{thermalgen}, a separate patch embedder tokenizes
$z_{\mathrm{vis}}$ and injects it into every Transformer block through
cross-attention, where the visible tokens serve as queries and the noised
latent tokens as keys and values. Working in the VAE latent space keeps the
XL/2 Transformer affordable while still processing full-resolution aerial
scenes, and flow matching turns sampling into a stable ODE solve even with
few steps. The backbone thus
inherits a stable infrared texture prior from pretraining, while its
condition pathway stays open for the physical factors introduced next.
\subsection{Physical Condition Design}
\label{sec:cond}
The thermal state of a land surface depends on the incoming solar
radiation and its heating and cooling history. On this basis, we
condition the generator on three factors computable from each frame's
timestamp and GPS coordinates, namely the solar elevation, the thermal phase clock,
and the flight altitude.
The solar elevation $\theta_e$ approximates the instantaneous irradiance
geometry and reflects the shortwave forcing of the surface energy balance at
the acquisition instant. However, the same elevation occurs in the morning and in the
afternoon, while the corresponding surface states differ because of the
accumulated heating. We therefore introduce the thermal phase clock (TPC)
$t_c \in [0, 24)$. Let $h$ denote the local time in hours. The local
sunrise and sunset times $h_r$ and $h_s$ are obtained from a solar
position algorithm \cite{reda}, and the day length is $D = h_s - h_r$.
The TPC counts the heating and cooling hours elapsed since sunrise on a
continuous 24-hour axis and is defined as
\begin{equation}
t_c =
\begin{cases}
h - h_r, & h_r \le h \le h_s,\\[1mm]
D + (h - h_s), & h > h_s,\\[1mm]
D' + (h + 24 - h_s'), & h < h_r
\end{cases}
\label{eq:tpc}
\end{equation}
where $h_s'$ and $D'$ are the sunset time and the day length of the previous
day, so that the cooling phase continues smoothly across midnight.
Unlike the local hour code of \emph{Naive-1D}, which one-hot encodes the
hour of day into 24 classes, the TPC follows the seasonal drift of sunrise and sunset, and thus
encodes thermal history rather than clock time. Identical values of $t_c$
correspond to similar accumulated heating or cooling on any date, which keeps
the condition comparable across the acquisition months. The flight altitude
$h_a$ captures the imaging scale of the acquisition and modulates the
apparent thermal contrast and brightness range of the scene. The final condition vector is
\begin{equation}
\mathbf{c} = \big[\, \theta_e / 90^{\circ},\;
\sin(2\pi t_c / 24),\; \cos(2\pi t_c / 24),\; h_a / 500 \,\big]
\label{eq:cond}
\end{equation}
where we encode the periodic TPC with a sine--cosine
pair to remove the discontinuity at the 24-hour boundary and normalize all
entries to a comparable range.
We deliberately keep the condition minimal and auditable rather than stacking
redundant dimensions. Within the compact site, the solar azimuth varies by
less than $0.2^{\circ}$ across frames at the same instant. Every retained factor thus has a
distinct physical meaning. The condition forms a compact
yet physically grounded coordinate for multitemporal generation.
\subsection{Solar Conditional Flow Matching}
\label{sec:scfm}
The generative
objective is then to transport a Gaussian prior to the target IR latent under
the guidance of both the source image and the physical condition. Specifically,
we inject the condition vector by replacing the label embedding of SiT with a
lightweight condition embedder,
\begin{equation}
\mathbf{y} = W_2\,\phi(W_1\,\mathbf{c})
\end{equation}
where $\phi$ is the SiLU activation, and the two layers project the
four-dimensional condition vector to the label dimension of SiT. We add $\mathbf{y}$ to the flow time
step embedding and feed it to every Transformer block through adaptive layer
normalization (adaLN) \cite{dit}. In each block, the scale and shift predicted
from $\mathbf{y}$ modulate the normalized activations, so the condition
adjusts the feature statistics globally while leaving the token sequence and
the attention pattern unchanged. The scale and shift are zero-initialized, so fine-tuning departs
smoothly from the pretrained modulation. To enable classifier-free guidance,
we replace $\mathbf{y}$ with a learnable null embedding at a probability of
$0.1$ during training. The network thereby learns both the conditional and
the unconditional velocity within a single set of weights.
Since registration black borders and orthophoto reprojection margins should
not dominate the supervision, the backbone is optimized with a masked flow matching
loss
\begin{equation}
\mathcal{L}_{\mathrm{fm}} =
\frac{\sum m \cdot \lVert v_\theta(z_t, t, z_{\mathrm{vis}}, \mathbf{c})
- v^{*}(z_t, t) \rVert_2^2}{\sum m}
\label{eq:loss}
\end{equation}
where $m$ is the valid-region mask downsampled to the latent resolution.
Only the SiT backbone and the condition embedder are optimized; both VAEs
remain frozen throughout, which preserves the pretrained generative prior and
keeps the trainable overhead small. At inference, a fifth-order
Dormand--Prince (dopri5) ODE solver transports Gaussian noise to the IR
latent with a budget of 50 steps, which balances fidelity and computational
cost. The thermal VAE then decodes the latent
into the final image. Classifier-free guidance combines the unconditional and
conditional velocities,
\begin{equation}
\tilde{v} = v_\theta(\cdot, \varnothing)
+ s\,\big( v_\theta(\cdot, \mathbf{c}) - v_\theta(\cdot, \varnothing) \big)
\label{eq:cfg}
\end{equation}
with a guidance scale $s = 4.0$, fixed for all experiments ($s = 1.0$ for \emph{Uncond}, whose conditional
branch is never trained). Crucially, the condition is computable
at any timestamp. A single visible image can thus be translated into IR
images at multiple times of the day, including hours never seen in
training. The initial noise is
fixed across conditions, and the generated differences therefore stem only
from the physical condition.
\begin{figure*}[!t]
\centering
\renewcommand{\arraystretch}{0.4}
\setlength{\tabcolsep}{1.2pt}
\newcommand{\fr}[1]{\includegraphics[width=0.12\textwidth]{#1}}
\newcommand{\rfr}[1]{\begingroup\setlength{\fboxsep}{1pt}\setlength{\fboxrule}{0.8pt}\fcolorbox{blue}{white}{\includegraphics[width=0.12\textwidth]{#1}}\endgroup}
\begin{tabular}{@{}c@{\hspace{3pt}}cc@{\hspace{8pt}}ccccc@{}}
 & Input & GT & 03:00 & 07:00 & 09:00 & 15:00 & 21:30 \\[1pt]
\raisebox{0.026\textwidth}{\rotatebox{90}{\footnotesize GT~15:22}}
 & \fr{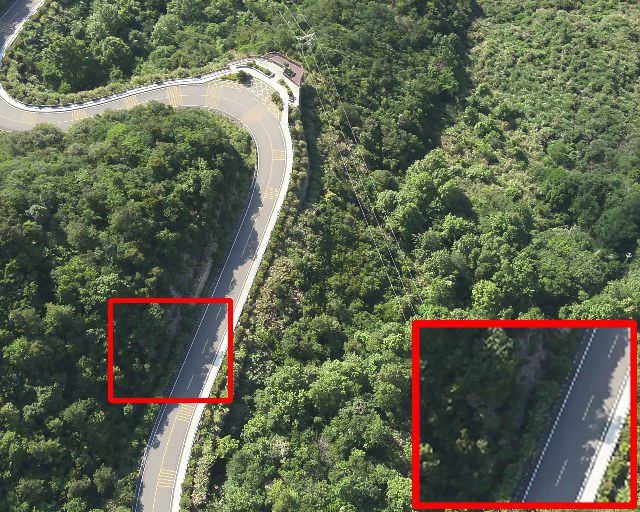}
 & \rfr{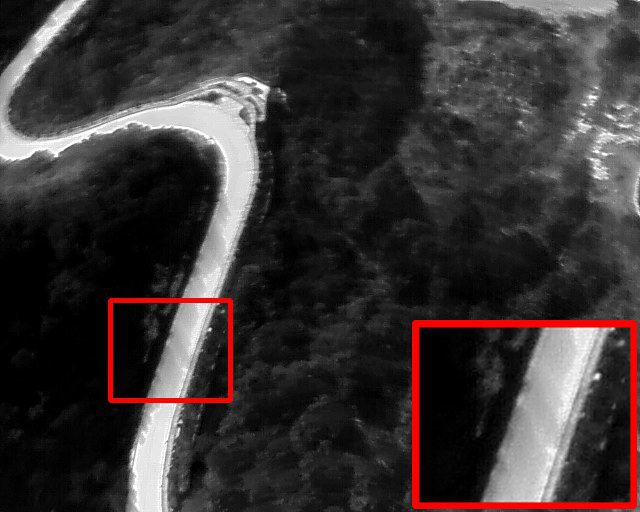}
 & \fr{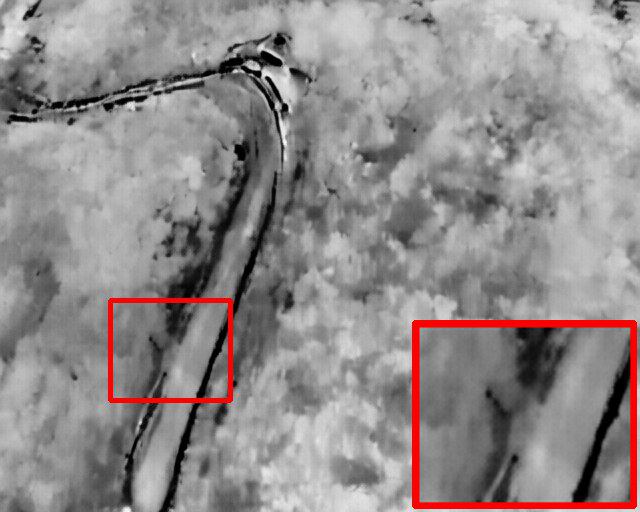}
 & \fr{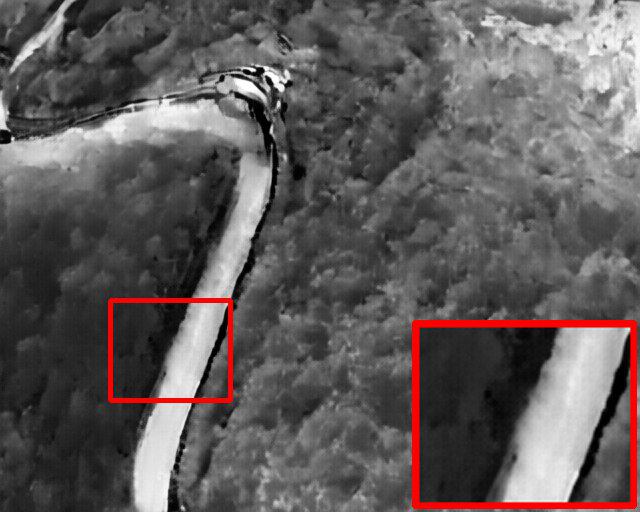}
 & \fr{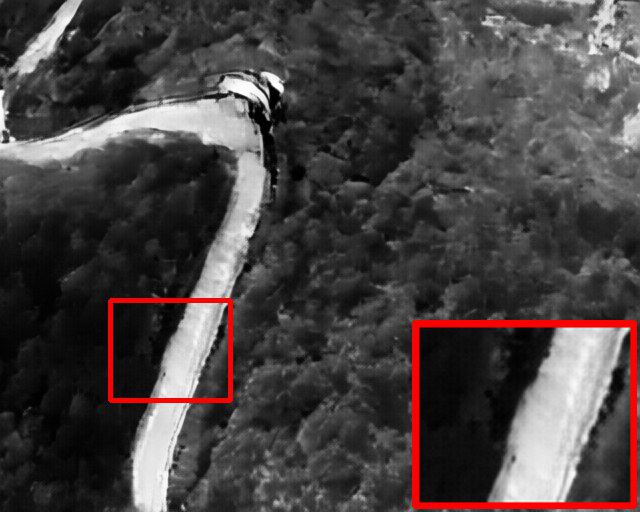}
 & \rfr{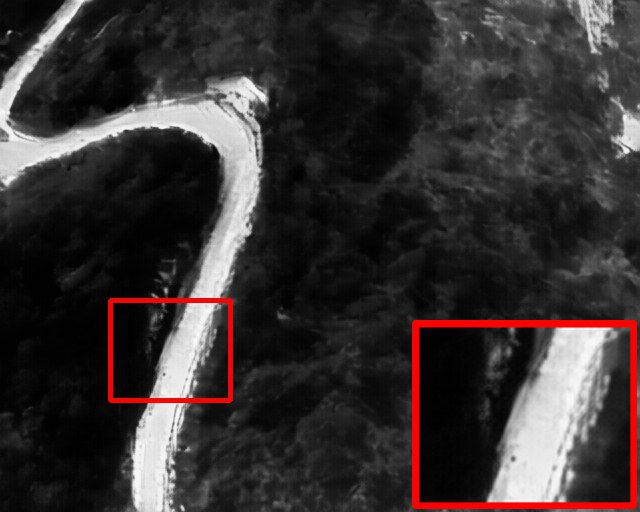}
 & \fr{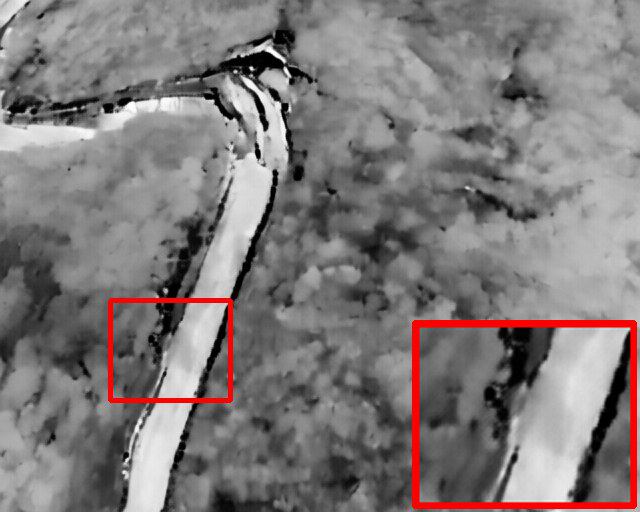} \\
\raisebox{0.026\textwidth}{\rotatebox{90}{\footnotesize GT~05:55}}
 & \fr{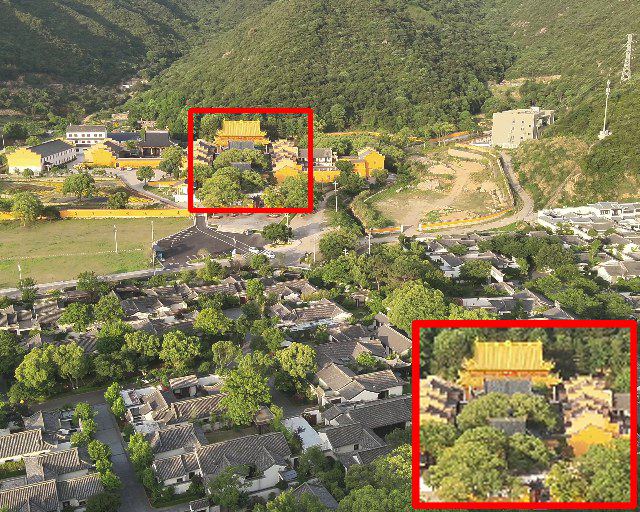}
 & \rfr{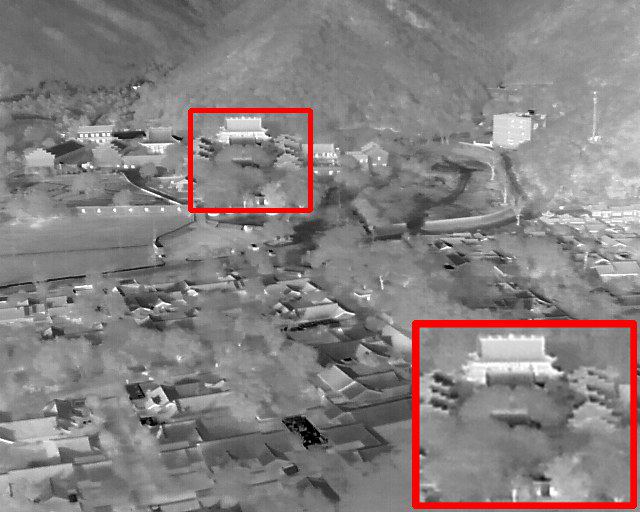}
 & \fr{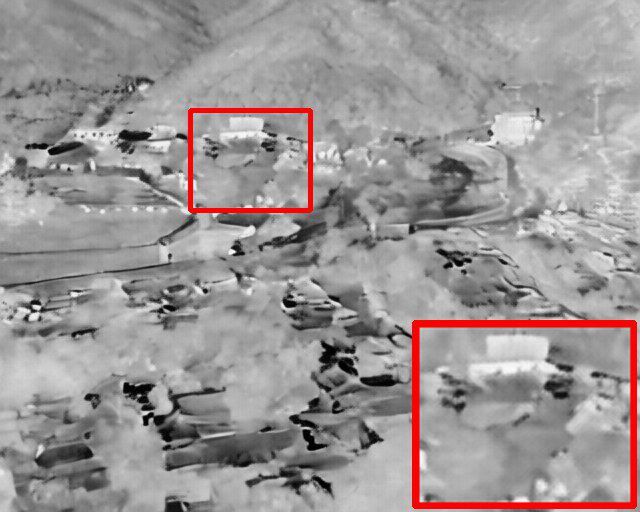}
 & \rfr{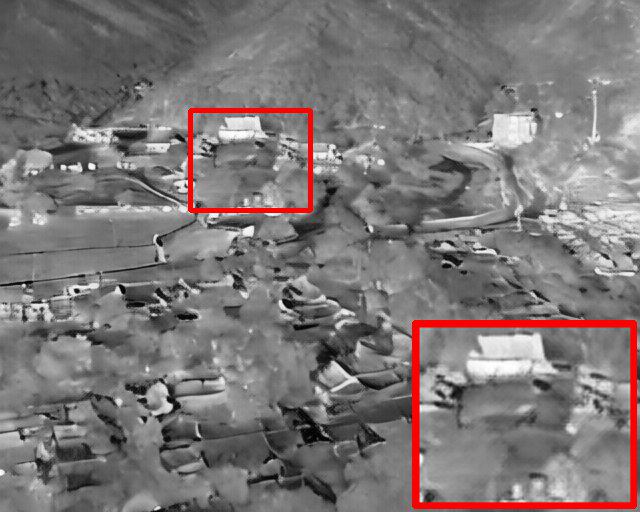}
 & \fr{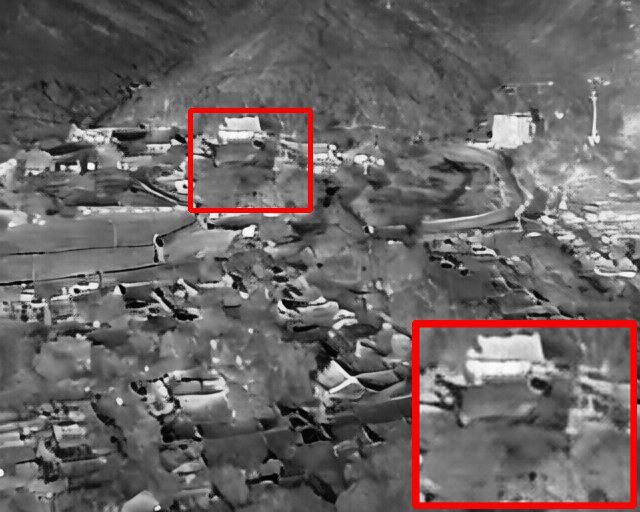}
 & \fr{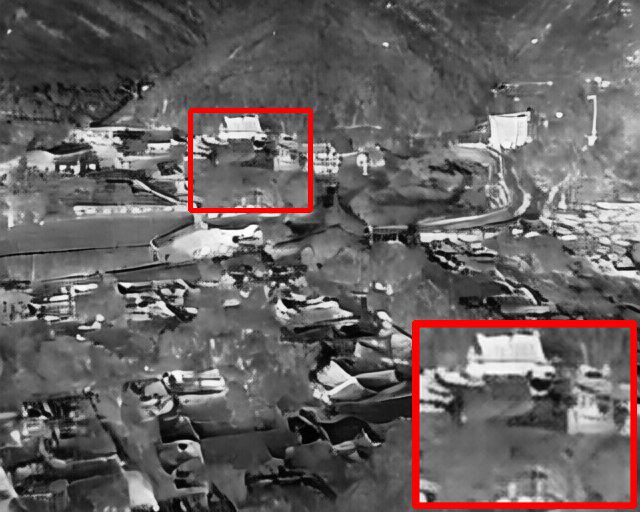}
 & \fr{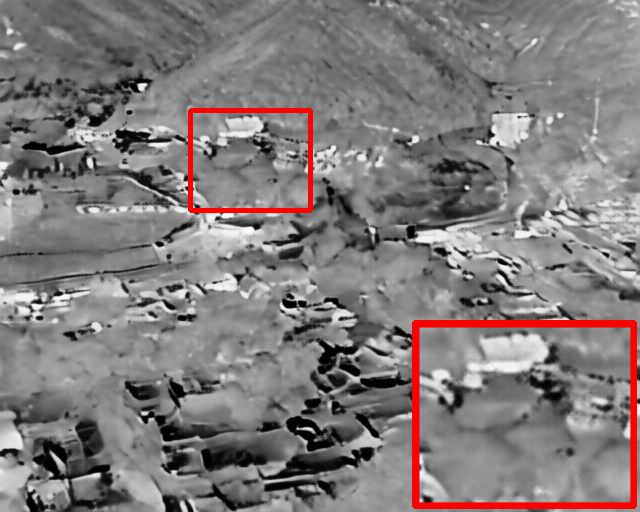} \\
\raisebox{0.026\textwidth}{\rotatebox{90}{\footnotesize GT~02:29}}
 & \fr{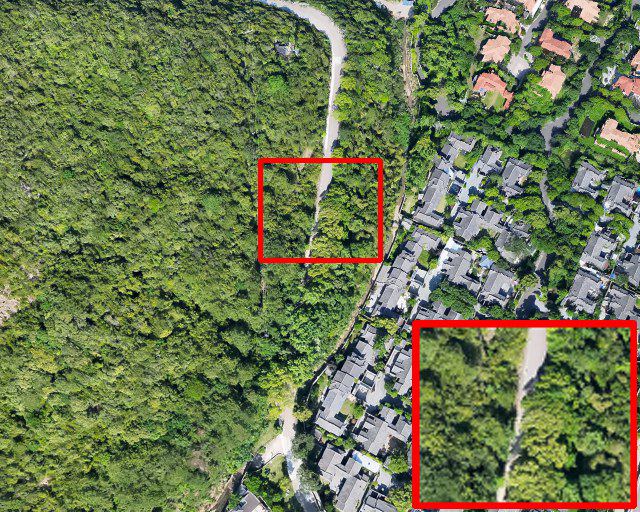}
 & \rfr{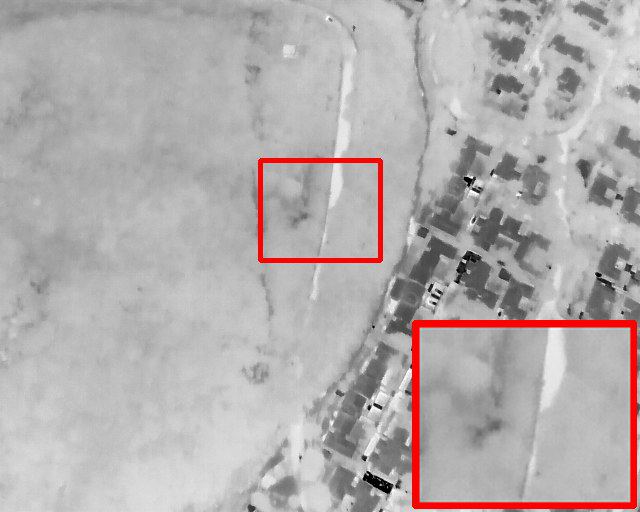}
 & \rfr{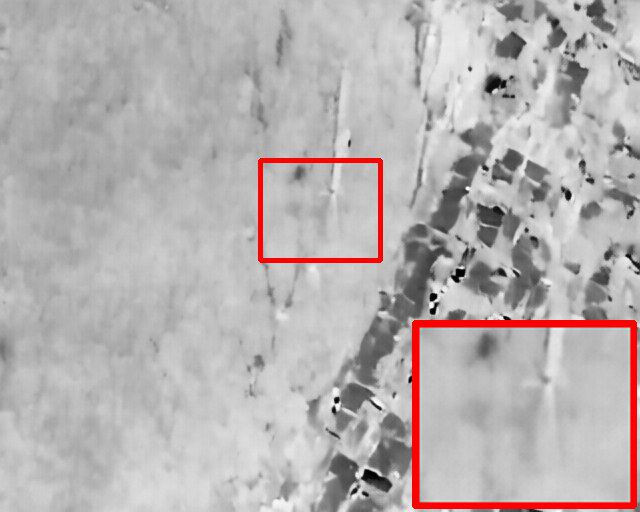}
 & \fr{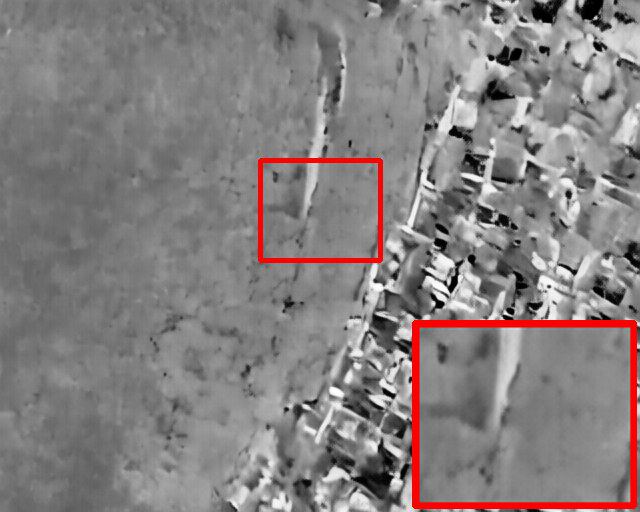}
 & \fr{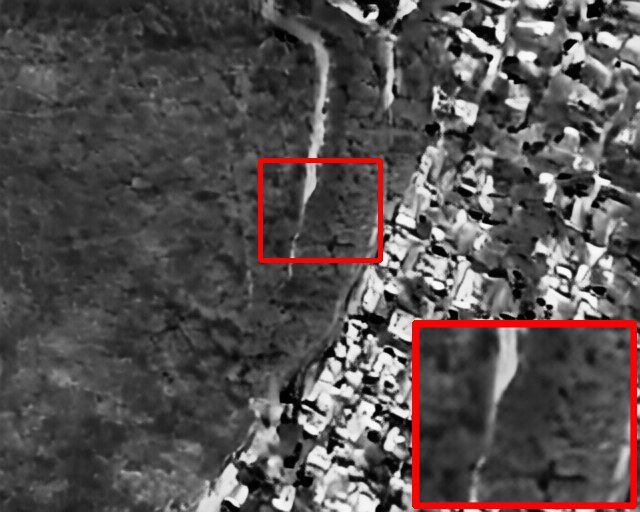}
 & \fr{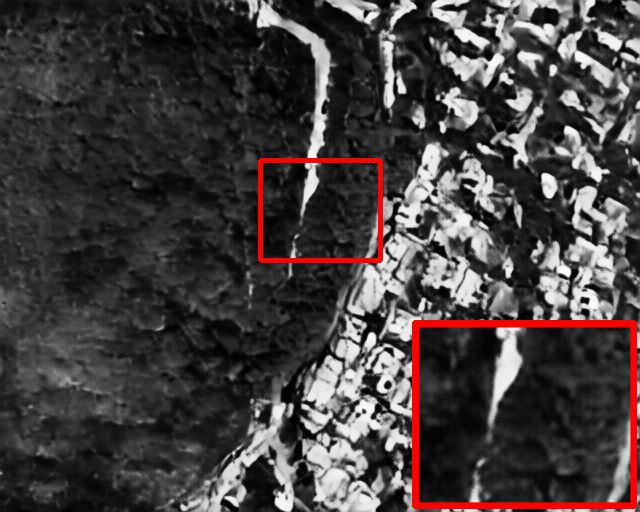}
 & \fr{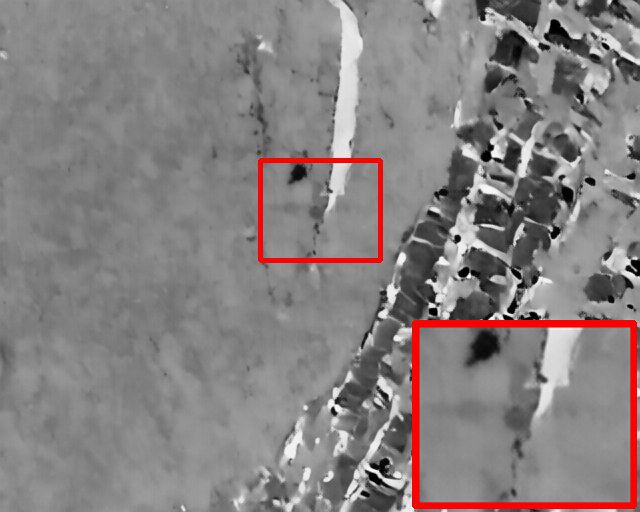} \\
\end{tabular}
\caption{Multitemporal generation from a single source image with fixed initial
noise. The leftmost label of each row is the acquisition time of the real GT
capture, which falls near one of the generated hours (GT~15:22 vs.\ 15:00,
GT~05:55 vs.\ 07:00, and GT~02:29 vs.\ 03:00). Blue frames mark the GT and the
temporally closest generated column, providing a same-scene real reference for
that row. Red boxes mark representative regions for detail inspection.}
\label{fig:multitime}
\end{figure*}
\section{Experiments and Analysis}
\subsection{Experimental Setup}
The dataset is collected with a DJI M3T UAV carrying synchronized visible
and thermal cameras in four acquisition
windows (morning 05:56--09:37, daytime 12:37--16:36, evening 19:56--21:30,
and night 02:29--03:46). Daytime pairs are registered from dual-camera frames with a
calibrated homography, whereas night pairs are built by reprojecting a
georeferenced orthophoto into each thermal frame. In total, the dataset
contains 5274 registered pairs in 121 sequences spanning 14 distinct hour
groups, with 4483 for training and 791 for testing. The split is made at the
sequence level, so frames of one flight sequence never appear in both sets.
The 5274 pairs comprise 1182, 2172, 1179, and 741 pairs in the morning,
daytime, evening, and night periods. Existing paired aerial
V2IR datasets cover only limited hours (Table~\ref{tab:dataset}).
We evaluate PSNR, SSIM, LPIPS
\cite{lpips}, and FID \cite{fid}, together with an unseen-hour error defined
in Section~\ref{sec:unseen}. All models are implemented via PyTorch and trained on a single NVIDIA
RTX A6000 GPU using the AdamW optimizer at a learning rate of
$5 \times 10^{-5}$ without weight decay. The full model contains about 675M parameters.
\begin{table}[H]
\caption{Comparison With Existing Paired Aerial V2IR Datasets}
\label{tab:dataset}
\centering
\footnotesize
\setlength{\tabcolsep}{2.5pt}
\begin{tabular}{ccc}
\toprule
Dataset & Task & Time coverage \\
\midrule
AVIID \cite{aviid} & V2IR translation & Daytime (2--5 p.m.) \\
DroneVehicle \cite{dronevehicle} & Detection & Day/night \\
CART \cite{cart} & Seg./trans. & Day/night \\
\textbf{Ours} & \textbf{Multitemporal V2IR trans.} & \textbf{14 hour groups} \\
\bottomrule
\end{tabular}
\end{table}
We compare different conditioning strategies on a
shared pretrained backbone, which isolates the effect of the condition design
from backbone capacity. All flow matching models are initialized from the pretrained
ThermalGen weights with both VAEs frozen.
We further train two baselines to validate
the condition design. In
\emph{Uncond}, the condition is always replaced by the null embedding,
which is equivalent to directly fine-tuning ThermalGen. In
\emph{Naive-1D}, the condition is the 24-class local-hour encoding defined
in Section~\ref{sec:cond}, fed through the same embedder. The Solar-Flow condition is also injected
into Pix2Pix
\cite{pix2pix} and CycleGAN \cite{cyclegan}, both retrained on the
proposed dataset. All variants follow the identical training schedule, so
their performance gaps arise solely from the condition design.
\subsection{Quantitative Comparison by Acquisition Period}
Table~\ref{tab:main} reports the quantitative comparison on the test frames
in the four acquisition periods. Solar-Flow matches \emph{Uncond} within
$2\%$ on PSNR and LPIPS, indicating that the
multitemporal control incurs almost no loss in translation fidelity.
The \emph{Naive-1D} ablation yields no consistent gain on these metrics,
because its scalar hour code maps different thermal states to the same value.
Among the three arms, Ours attains the best FID ($96.12$).
All arms reach their lowest
PSNR in the daytime period, where strong solar loading produces the largest
thermal contrast and the hardest regression. Conversely, the compressed
thermal range at night yields the highest SSIM. These per-frame
metrics barely separate the three arms; the decisive differences lie in the
FID and unseen-hour error columns. Fidelity at the captured hours is only
a prerequisite, and the next three experiments reveal its real value.
\begin{table}[!t]
\caption{Quantitative comparison by acquisition period. All metrics are
computed on the same test frames.}
\label{tab:main}
\centering
\footnotesize
\setlength{\tabcolsep}{3.5pt}
\begin{tabular}{ccccccc}
\toprule
Method & Period & PSNR$\uparrow$ & SSIM$\uparrow$ & LPIPS$\downarrow$ & FID$\downarrow$ & Unseen Err$\downarrow$ \\
\midrule
\multirow{4}{*}{\emph{Uncond}}
 & Morning & 20.92 & 51.18\% & 44.74\% & \multirow{4}{*}{99.72} & \multirow{4}{*}{34.7} \\
 & Daytime & 18.33 & 44.81\% & 46.90\% & & \\
 & Evening & \textbf{19.03} & 56.41\% & \textbf{48.96\%} & & \\
 & Night   & \textbf{19.95} & 61.73\% & \textbf{46.79\%} & & \\
\midrule
\multirow{4}{*}{\emph{Naive-1D}}
 & Morning & 20.53 & 50.51\% & 45.23\% & \multirow{4}{*}{102.05} & \multirow{4}{*}{25.1} \\
 & Daytime & 18.03 & 44.15\% & 47.46\% & & \\
 & Evening & 18.85 & \textbf{57.19\%} & 50.10\% & & \\
 & Night   & 19.72 & \textbf{62.47\%} & 48.68\% & & \\
\midrule
\multirow{4}{*}{\textbf{Ours}}
 & Morning & \textbf{20.95} & \textbf{51.40\%} & \textbf{44.02\%} & \multirow{4}{*}{\textbf{96.12}} & \multirow{4}{*}{\textbf{15.7}} \\
 & Daytime & \textbf{18.41} & \textbf{45.24\%} & \textbf{46.53\%} & & \\
 & Evening & 19.01 & 57.09\% & 49.25\% & & \\
 & Night   & 19.89 & 61.93\% & 47.04\% & & \\
\bottomrule
\end{tabular}
\end{table}
\subsection{Multitemporal Generation}
To verify that the model learns the multitemporal mapping rather than a
fixed one, Fig.~\ref{fig:multitime} translates one source image
under five conditions (03:00, 07:00, 09:00, 15:00, and 21:30)
with the initial noise fixed. Across these conditions, the generated images
preserve the scene structure of the input, while brightness and land-cover
contrast evolve consistently with the thermal states. For
example, the scenes are brightest in the early morning and darkest around
noon, and the relative contrast among roofs, roads, and vegetation changes
consistently across the day.
The zoomed insets further confirm that fine
structures, such as road markings and rooftop edges, remain sharp and aligned
across all hours. Since the columns of each row share the initial noise, all
differences between them originate from the physical condition alone,
indicating that Solar-Flow separates the scene structure from the thermal
state.
\subsection{Generalization to Unseen Hours}
\label{sec:unseen}
\begin{figure}[H]
\centering
\setlength{\tabcolsep}{1pt}
\begin{tabular}{@{}c c@{\hspace{8pt}}c c c c@{}}
 & & & {\scriptsize 03:00} & {\scriptsize \textbf{04:30}$^*$ (unseen)} & {\scriptsize 05:00} \\[1pt]
\multirow{2}{*}{\begin{tabular}{@{}c@{}}\includegraphics[width=0.115\linewidth]{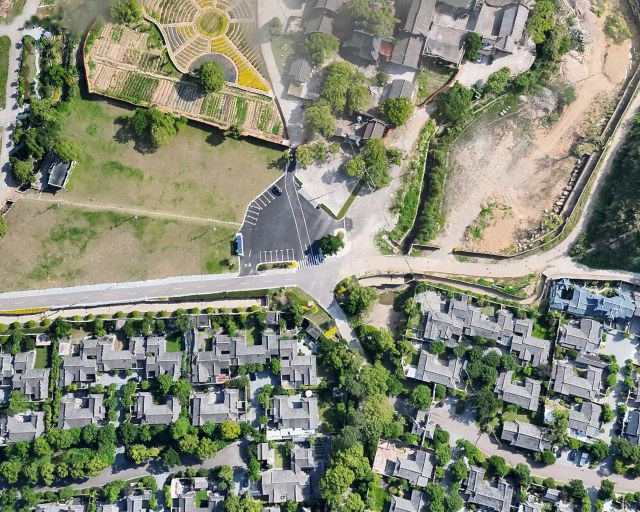}\\[1.5pt]{\scriptsize Input}\end{tabular}} &
\multirow{2}{*}{\begin{tabular}{@{}c@{}}\includegraphics[width=0.115\linewidth]{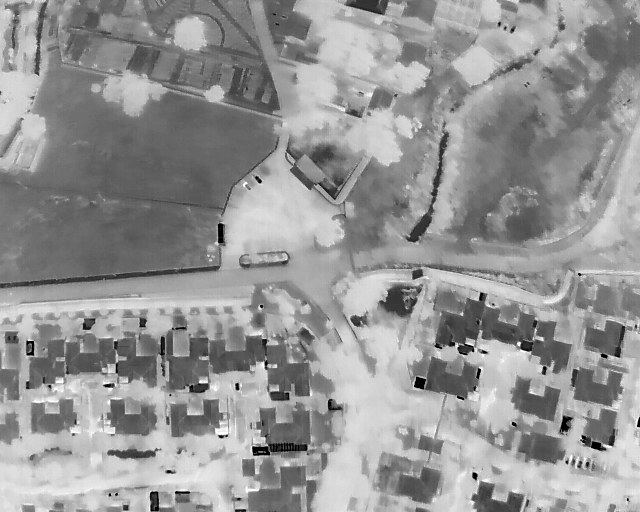}\\[1.5pt]{\scriptsize GT (03:44)}\end{tabular}} &
\raisebox{0.042\linewidth}{\rotatebox{90}{\scriptsize Ours}} &
\includegraphics[width=0.185\linewidth]{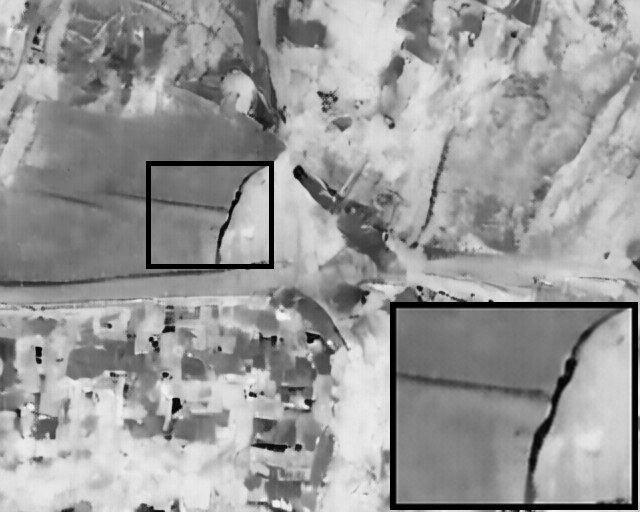} &
\includegraphics[width=0.185\linewidth]{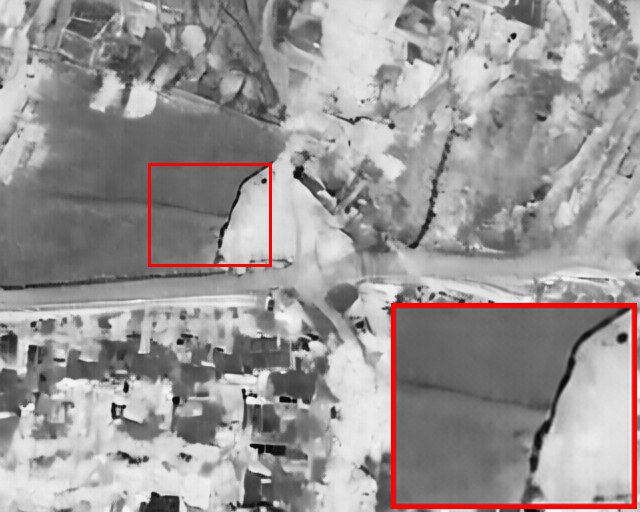} &
\includegraphics[width=0.185\linewidth]{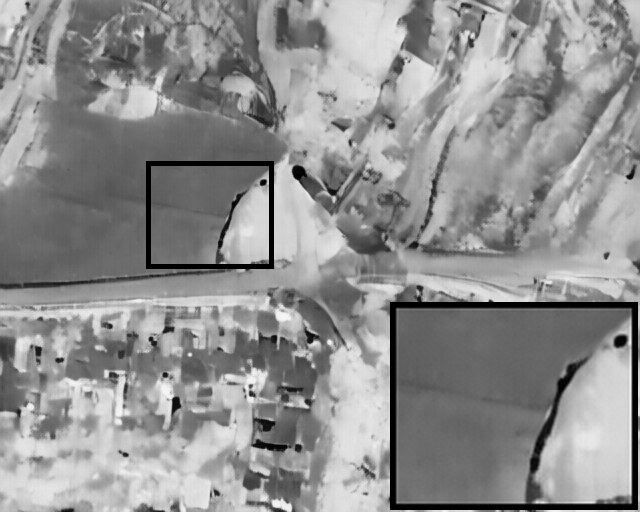} \\
 & &
\raisebox{0.042\linewidth}{\rotatebox{90}{\scriptsize Naive-1D}} &
\includegraphics[width=0.185\linewidth]{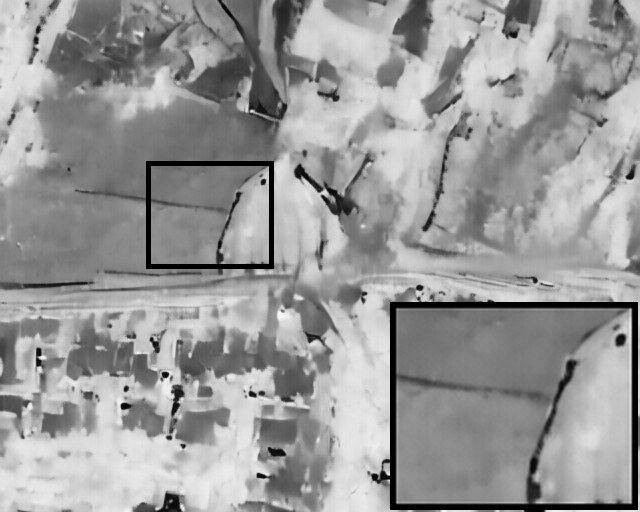} &
\includegraphics[width=0.185\linewidth]{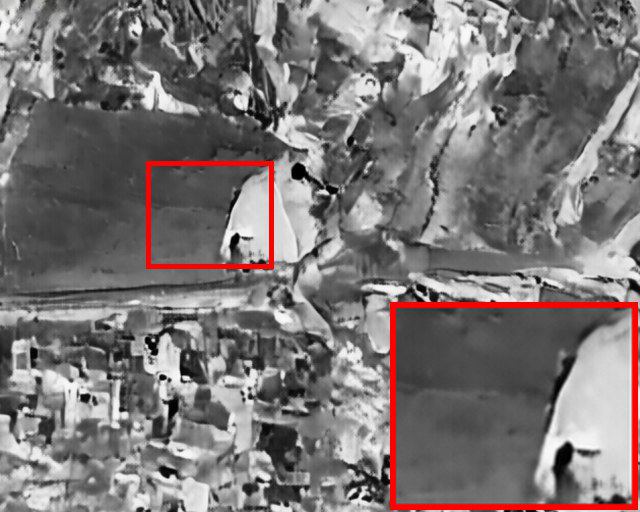} &
\includegraphics[width=0.185\linewidth]{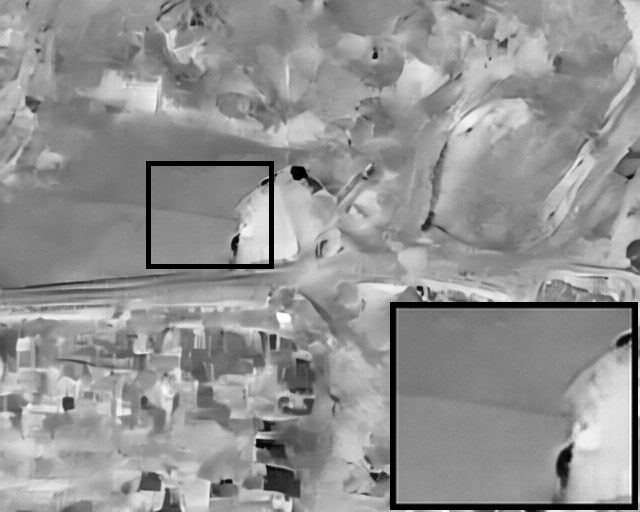} \\
\end{tabular}%
\caption{Extrapolation to the unseen pre-dawn hour 04:30$^*$, with a real
capture at 03:44 as the reference. The continuous physical condition (Ours)
yields a thermal state close to the real capture.}
\label{fig:dusk}
\end{figure}
A more demanding test concerns hours entirely absent from training: we
directly generate an IR image at 04:30 with the trained model
(Fig.~\ref{fig:dusk}). Since the
TPC and the solar elevation remain continuous and physically defined at this
hour, the generated appearance transitions smoothly toward the
corresponding thermal state and closely matches the real capture
at 03:44. The \emph{Naive-1D} baseline instead extrapolates an embedding value
never seen during training and produces a degraded, physically inconsistent
appearance. The unseen-hour error in Table~\ref{tab:main} confirms this
advantage, defined as the mean-brightness deviation from the interpolated
ground-truth trend at hours never captured in training. Ours reaches $15.7$
against $25.1$ for \emph{Naive-1D} and $34.7$ for \emph{Uncond}.
\subsection{Generalization to Other Architectures}
\begin{figure}[H]
\centering
\setlength{\tabcolsep}{1.5pt}
\renewcommand{\arraystretch}{1}
\begin{tabular}{cc@{\hspace{8pt}}c@{\hspace{1pt}}cccc}
 & & & {\scriptsize 03:00} & {\scriptsize 15:00} & {\scriptsize 21:30} \\[1pt]
\multirow{3}{*}[-14pt]{{\begin{tabular}{@{}c@{}}\raisebox{-0.015\linewidth}{\includegraphics[width=0.105\linewidth]{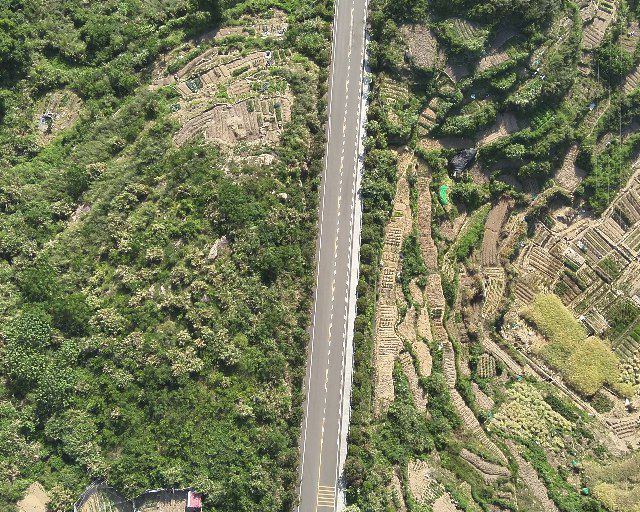}}\\[1.5pt]{\scriptsize Input}\end{tabular}}} &
\multirow{3}{*}[-14pt]{{\begin{tabular}{@{}c@{}}\raisebox{-0.015\linewidth}{\includegraphics[width=0.105\linewidth]{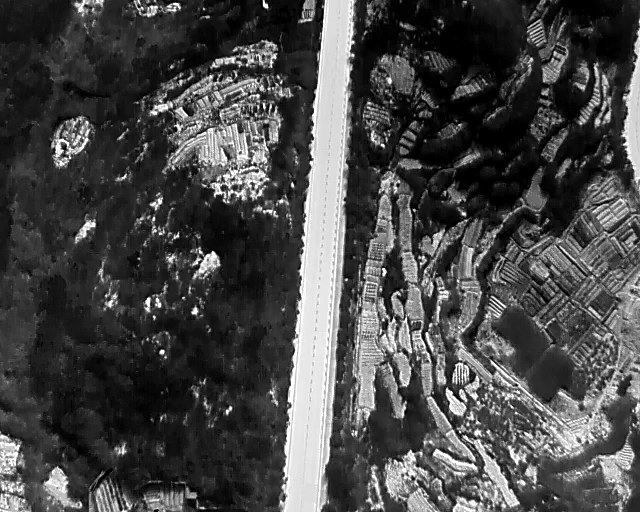}}\\[1.5pt]{\scriptsize GT 15:54}\end{tabular}}} &
\raisebox{0.012\linewidth}{\rotatebox{90}{\scriptsize \shortstack{Pix2Pix\\+cond}}} &
\includegraphics[width=0.17\linewidth]{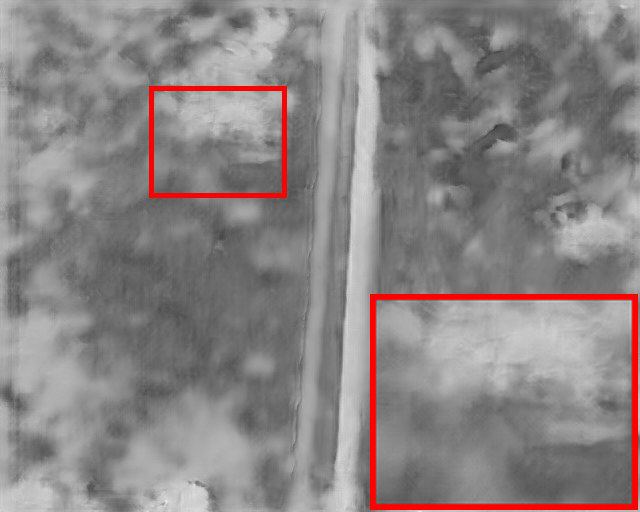} &
\includegraphics[width=0.17\linewidth]{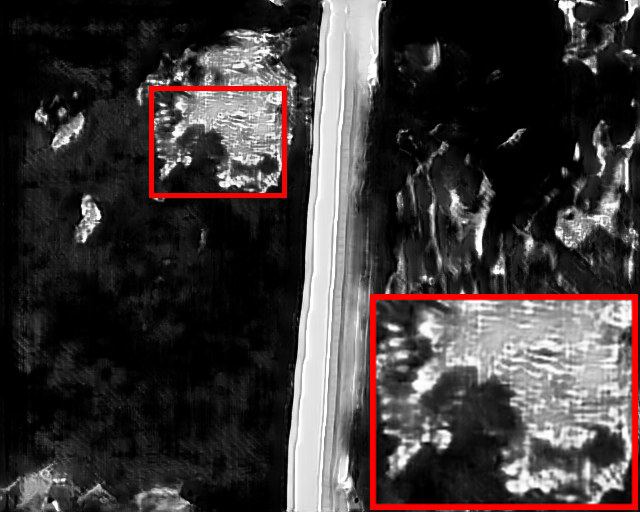} &
\includegraphics[width=0.17\linewidth]{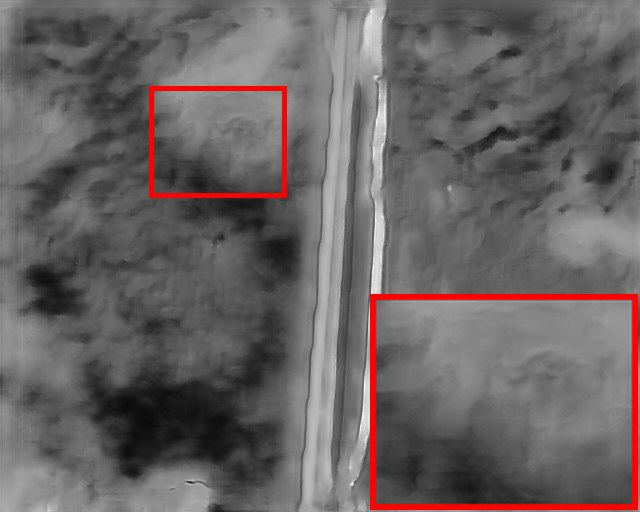} \\
 & &
\raisebox{0.012\linewidth}{\rotatebox{90}{\scriptsize \shortstack{CycleGAN\\+cond}}} &
\includegraphics[width=0.17\linewidth]{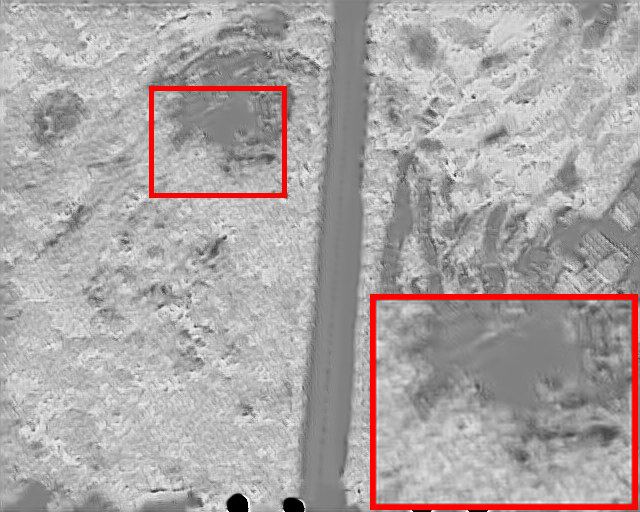} &
\includegraphics[width=0.17\linewidth]{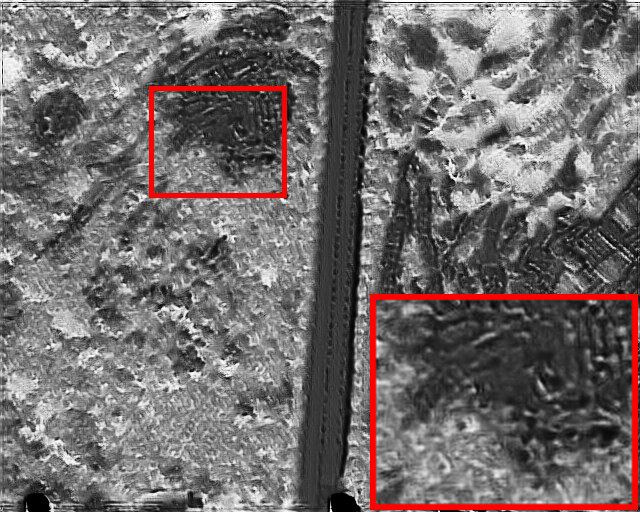} &
\includegraphics[width=0.17\linewidth]{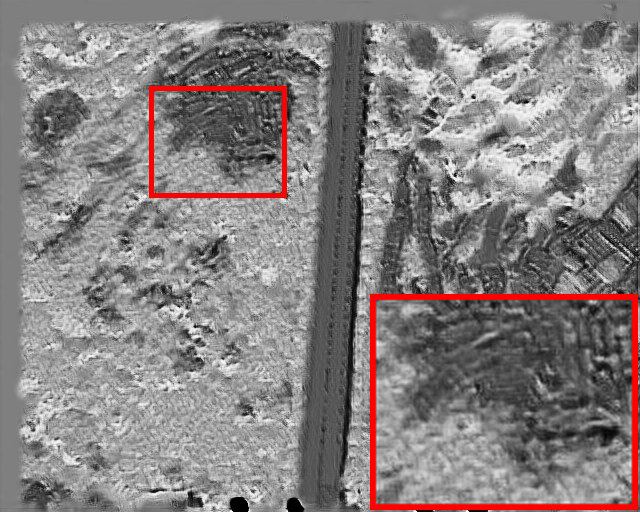} \\
 & &
\raisebox{0.024\linewidth}{\rotatebox{90}{\scriptsize \textbf{Ours}}} &
\includegraphics[width=0.17\linewidth]{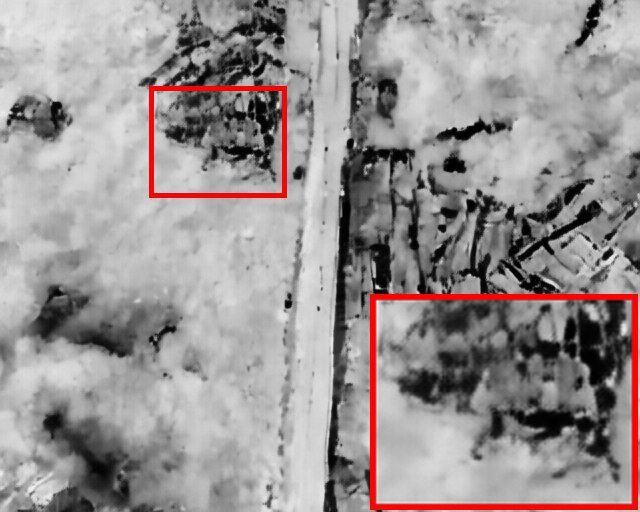} &
\includegraphics[width=0.17\linewidth]{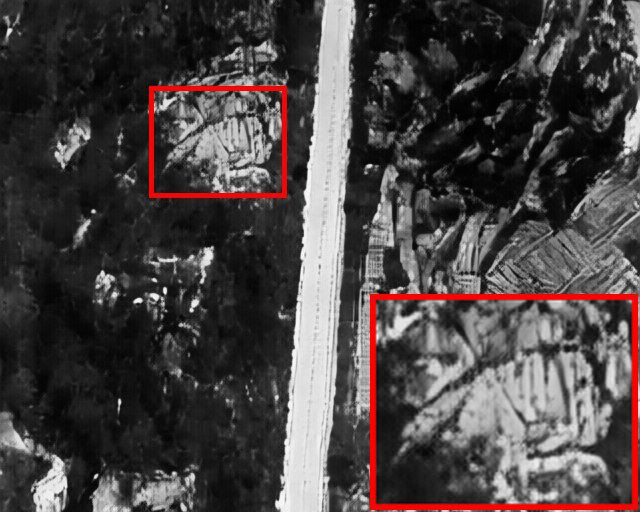} &
\includegraphics[width=0.17\linewidth]{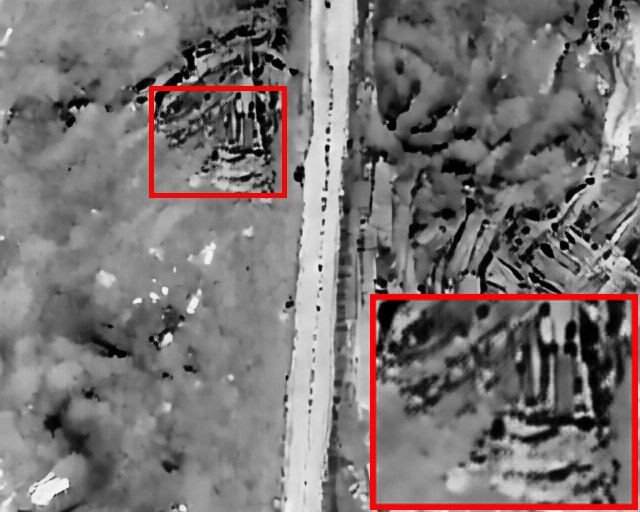} \\
\end{tabular}
\caption{Multitemporal generation of the translators and Ours.
The translators respond to the target hour but exhibit blurred textures and
artifacts, whereas Ours preserves sharp thermal details across
the day.}
\label{fig:gan}
\end{figure}

To answer whether the multitemporal capability stems from the physical
condition or from the flow matching backbone, Table~\ref{tab:gan} and
Fig.~\ref{fig:gan} compare the alternative translators before and after
injecting the proposed condition. After the injection, both Pix2Pix and
CycleGAN follow the target hour across the day, and the unseen-hour error
drops from $34.1$ to $21.6$ for Pix2Pix and from $34.5$ to $26.4$ for
CycleGAN.
Conversely, the retrained unconditional Pix2Pix collapses to a single
deterministic mapping and cannot separate the hours at all, despite an FID
of $127.31$. These results confirm that the physical condition grants
translators the multitemporal capability, not the backbone.
However, this capability alone does not deliver generation quality, which
still relies on the backbone. As Fig.~\ref{fig:gan} shows, the conditioned GANs
produce blurred textures and distorted structures. They lag behind Ours on
all perceptual metrics. Their LPIPS rises to $61.82\%$ and $63.26\%$ versus
$46.56\%$ for Ours, and their FID rises to $219.43$ and $188.76$ versus
$96.12$. Moreover, CycleGAN+cond exhibits radiometric drift in late training
and requires checkpoint selection on the validation set. Both conditioned
GANs also degrade beyond the captured hours, whereas the flow matching model
handles the same condition correctly (Fig.~\ref{fig:dusk}). Thus, the
physical condition enables multitemporal generation, while the flow
matching backbone remains essential for the generation quality and
extrapolation stability.
\begin{table}[H]
\caption{Effect of the proposed condition on alternative translators.}
\label{tab:gan}
\centering
\footnotesize
\setlength{\tabcolsep}{4pt}
\begin{tabular}{cccccc}
\toprule
Method & PSNR$\uparrow$ & SSIM$\uparrow$ & LPIPS$\downarrow$ & FID$\downarrow$ & Unseen Err$\downarrow$ \\
\midrule
Pix2Pix             & 19.03 & 47.11\% & 51.27\% & 127.31 & 34.1 \\
Pix2Pix+cond        & 17.06 & 48.02\% & 61.82\% & 219.43 & 21.6 \\
CycleGAN            & 15.41 & 45.58\% & 61.50\% & 128.29 & 34.5 \\
CycleGAN+cond       & 15.21 & 42.24\% & 63.26\% & 188.76 & 26.4 \\
\textbf{Ours}       & \textbf{19.37} & \textbf{51.48\%} & \textbf{46.56\%} & \textbf{96.12} & \textbf{15.7} \\
\bottomrule
\end{tabular}
\end{table}
\section{Conclusion}
This letter presents Solar-Flow, which conditions a pretrained flow matching
backbone on three computable physical factors, namely the solar elevation, the
thermal phase clock, and the flight altitude. The framework upgrades aerial V2IR
translation from static mapping to multitemporal generation, together with a
paired dataset covering 14 hour groups. Experiments on the test subset confirm that the physical
condition preserves translation fidelity on par with the \emph{Uncond} and
\emph{Naive-1D} baselines while extrapolating plausibly to unseen hours. Future
work will extend the current single-site dataset to multiple sites and seasons
with radiometric calibration and meteorological records.

\end{document}